\documentclass{Interspeech2024}

\interspeechcameraready

\title{Zero-Shot End-To-End Spoken Question Answering In Medical Domain}

\name[affiliation={1,2}]{Yanis}{Labrak}
\name[affiliation={1}]{Adel}{Moumen}
\name[affiliation={1,3}]{Richard}{Dufour}
\name[affiliation={1}]{Mickael}{Rouvier}

\address{
  $^1$LIA - Avignon University, France \hspace{3mm} $^2$Zenidoc, France \\
  $^3$Nantes Université, École Centrale Nantes, CNRS, LS2N, UMR 6004, F-44000 Nantes, France}
\email{yanis.labrak@univ-avignon.fr, adel.moumen@univ-avignon.fr, richard.dufour@univ-nantes.fr, mickael.rouvier@univ-avignon.fr}


\usepackage{graphicx}
\usepackage{multirow}
\usepackage{arydshln}
\usepackage{float}

\usepackage[T1]{fontenc}

\usepackage[skins]{tcolorbox}

\definecolor{myblue}{RGB}{28, 144, 153}
\definecolor{mybluelight}{RGB}{236, 226, 240}

\newtcolorbox{instructionframe}[2][]{%
  enhanced,colback=white,colframe=myblue,coltitle=white,boxrule=1.0pt,
  fonttitle=\mdseries,
  attach boxed title to top left={yshift=-0.5\baselineskip-0.4pt,xshift=2mm},
  boxed title style={tile,size=minimal,left=1.5mm,right=1.5mm,
    colback=myblue,before upper=\strut},
  title=#2,#1
}

\begin{document}

\maketitle

\begin{abstract}

In the rapidly evolving landscape of spoken question-answering (SQA), the integration of large language models (LLMs) has emerged as a transformative development. Conventional approaches often entail the use of separate models for question audio transcription and answer selection, resulting in significant resource utilization and error accumulation. To tackle these challenges, we explore the effectiveness of end-to-end (E2E) methodologies for SQA in the medical domain. Our study introduces a novel zero-shot SQA approach, compared to traditional cascade systems. Through a comprehensive evaluation conducted on a new open benchmark of 8 medical tasks and 48 hours of synthetic audio, we demonstrate that our approach requires up to 14.7 times fewer resources than a combined 1.3B parameters LLM with a 1.55B parameters ASR model while improving average accuracy by 0.5\%. These findings underscore the potential of E2E methodologies for SQA in resource-constrained contexts.

\end{abstract}

\section{Introduction}

Spoken Question Answering (SQA) aims to identify the correct answer from spoken documents or texts in response to a given spoken query. Unlike many other spoken language understanding tasks, such as speech summarization, which primarily focus on semantic comprehension at the utterance level, SQA demands advanced comprehension and reasoning over extensive audio content. In addition to grasping the question and understanding the global context within the audio, it requires capturing nuanced details to accurately select the correct answer, often involving the utilization of out-of-context information. As a result, SQA poses a significant challenge due to its multifaceted nature.

Traditionally, SQA methods comprise a cascade of an Automatic Speech Recognition (ASR) system to transcribe the audio question followed by a Language Model (LM). The LM takes a prompt and the automatic transcription as input to predict the correct answer among a list of options. However, ASR errors introduce noise into the LM input, leading to performance degradation and information loss, despite community efforts~\cite{8683377, lee18d_interspeech, 9414999} to mitigate these issues and enhance robustness to transcription errors. Consequently, the cascade of stages cannot match the performance of a single-stage model based on speech due to inherent information loss~\cite{22eb39cdbd8f45b8bc46db3895615cd8}. 

The emergence of Large Language Models (LLMs) like Bloom~\cite{workshop2023bloom} or LLaMa 2~\cite{touvron2023llama} represents a significant advancement in question-answering systems. However, these models require extensive parameter scaling, further complicating the challenge of running separate models for each stage. For instance, the ASR models are already large (e.g., Whisper Medium with 769M parameters and Large V2 with 1.55B parameters), necessitating significant hardware resources for each stage. Consequently, there is a growing interest in directly extracting information from speech to preserve maximal information while minimizing hardware requirements.



Several architectures, such as Whisper~\cite{radford2022robust}, CLAP~\cite{laionclap2023}, and SpeechT5~\cite{ao-etal-2022-speecht5}, have proposed unifying textual and audio modalities using encoder-decoder models. Notably, autoregressive approaches based on LLMs, exemplified by SpeechGPT~\cite{zhang-etal-2023-speechgpt}, have emerged. These models rely on textual prompts to encode speech signals into discrete units.

We propose a novel end-to-end audio-text entailment strategy for zero-shot multiple-choice question answering tasks, focusing on the medical domain. Inspired by zero-shot classification methods in textual Natural Language Processing (NLP)~\cite{halder-etal-2020-task, pamies-etal-2023-weakly} and computer vision~\cite{du-etal-2023-zero, Mercea_2022_CVPR}, our approach leverages the model's capacity to identify modalities that entail each other. Our contributions include:

\begin{itemize}

\item An innovative audio-text entailment approach for zero-shot spoken multiple-choice question answering tasks.

\item A new SQA dataset tailored to the medical domain.

\item A zero-shot performance comparison of 4 existing state-of-the-art end-to-end models.

\item An in-depth analysis of the disposition of the information required for the SQA task within speech encoder layers.

\item A public release of all the code and data on GitHub and Hugging Face~\footnote{\href{https://huggingface.co/SpokenMedicalQA}{https://huggingface.co/SpokenMedicalQA}}.


\end{itemize}

\section{Medical spoken question answering}


In this section, we define the SQA task (Section~\ref{s:def}) and present the open benchmark constructed from established medical datasets initially in textual format (Section~\ref{s:tasks}). Additionally, we describe the audio prompt format (Section~\ref{s:prompt}) and the SQA evaluation protocol (Section~\ref{s:eval}).


\subsection{Definition}
\label{s:def}

We focus on multiple-choice SQA within the medical domain. Each instance comprises an audio question followed by four possible spoken responses, denoted as $(q, o, c, a)$. Here, $q$ represents the question, $o$ denotes the options (labeled A to D), $c$ indicates the correct answer and $a$ encapsulates the audio containing both the question and options. Questions are structured as single-turn interactions, devoid of dialogue.
This evaluation relies solely on the model's internal knowledge without external information or span extraction. The primary objective is to assess end-to-end model performance in understanding and accurately choosing the correct answer from spoken input. 






\subsection{Tasks collection and description}
\label{s:tasks}









Recent years have seen significant progress in SQA datasets, such as Clotho-AQA~\cite{lipping2022clothoaqa}, Spoken-SQuAD~\cite{lee2018spoken}, and LibriSQA~\cite{zhao2023librisqa}. However, these datasets do not specifically target the healthcare domain or rely solely on audio inputs. The absence of SQA datasets in the medical domain hampers the development of question answering systems tailored to healthcare contexts. To address this gap, we propose synthesizing an audio dataset from existing textual multiple-choice question answering (MCQA) corpora. Our approach involves using Text-To-Speech (TTS) technology on these MCQA textual datasets to generate synthetic audios, leveraging advancements in TTS models that increasingly resemble human speech quality~\cite{pmlr-v139-kim21f, 10.5555/3495724.3497152}. We utilized the OpenAI TTS API (\texttt{tts-1}) to synthesize speech based on the questions and available options. The speakers were alternated through the 6 available voices to introduce diversity and realism into the dataset. The resulting audio files were sampled at 16,000 Hz and converted to WAV mono channel format.

Our reference texts were sourced from three open-source textual MCQA corpora in English, all relevant to healthcare, featuring single possible answers and a four-option format. Note that only the test data are detailed here, as the proposed approaches operate under zero-shot conditions.


%

\textbf{MMLU}~\cite{hendrycks2021measuring} comprises exam questions spanning 57 subjects, including those relevant to healthcare. We focused on six healthcare-related subjects already evaluated in MedPaLM-2~\cite{singhal2023expertlevel}: college biology, college medicine, anatomy, professional medicine, medical genetics, and clinical knowledge. The dataset includes a test set of 1,089 questions, totaling 8 hours and 39 minutes of synthesized audio.

\textbf{MedQA}~\cite{jin2020disease} integrates questions formatted similarly to the US Medical License Exam (USMLE), covering diverse medical topics. We exclusively utilized the test set, comprising 1,273 questions amounting to 21 hours and 22 minutes of audio.


\textbf{MedMCQA}~\cite{pmlr-v174-pal22a} consists of questions with four options each, extracted from Indian medical entrance examinations (AIIMS/NEET). It encompasses 2,400 healthcare topics across 21 medical subjects, with 4,183 questions for the validation which are used as test ones since it is unavailable to the public~\cite{wu2023pmcllama}. The test set comprises 17 hours and 40 minutes of audio.


Our final benchmark encompasses 8 SQA tasks (including 6 from MMLU) derived from these 3 synthesized datasets. Table~\ref{tab:audio-distribution} summarizes the audio duration distribution according to the different labels available in the test set.

\begin{table*}[!t]
\centering
\setlength\extrarowheight{1.5pt}
\caption{Accuracy (in \%) of the zero-shot cascade methods. 
Highest value in bold and second best is underlined.}
\label{tab:results-cascade}
\resizebox{\textwidth}{!}{%

\begin{tabular}{lllllllllll}
\hline

\multicolumn{1}{c}{\textbf{}} &
\multicolumn{1}{c}{\textbf{}} &
  \multicolumn{6}{c}{\textbf{MMLU}} &
  \multicolumn{1}{c}{\textbf{}} &
  \multicolumn{1}{c}{\textbf{}} \\ \cline{2-8}
  
\multicolumn{1}{c}{\textbf{}} &
\multicolumn{1}{c}{\textbf{}} &
  \multicolumn{1}{c}{\textbf{Clinical KG}} &
  \multicolumn{1}{c}{\textbf{Medical Genetics}} &
  \multicolumn{1}{c}{\textbf{Anatomy}} &
  \multicolumn{1}{c}{\textbf{Pro Medicine}} &
  \multicolumn{1}{c}{\textbf{College Biology}} &
  \multicolumn{1}{c}{\textbf{College Medicine}} &
  \multicolumn{1}{c}{\textbf{MedQA}} &
  \multicolumn{1}{c}{\textbf{MedMCQA}} &
  \multicolumn{1}{c}{\textbf{Avg.}} \\
  \hline

\multirow{4}{*}{Phi 1.5} & Oracle &
31.3 & \textbf{39.0} & 19.3 & 20.6 & 29.2 & 28.9 & 27.7 & 31.2 &
28.4
\\
& Whisper Small &
26.8 & 24.0 & 31.9 & 27.6 & 25.0 & 23.1 & 25.5 & 25.9 &
26.2
\\
& Whisper Medium &
27.9 & 20.0 & 35.6 & 27.6 & 25.7 & 24.9 & 25.4 & 25.4 &
26.6
\\
& Whisper Large V2 &
31.7 & 19.0 & 34.1 & 24.6 & 26.4 & 26.0 & 27.6 & 26.2 &
27.0
\\
   \hdashline

\multirow{4}{*}{Llama 2 7B} & Oracle &
21.5 & 30.0 & 18.5 & 18.4 & 25.7 & 20.8 & 27.7 & 32.1 &
24.3
\\
 & Whisper Small &
29.4 & 31.0 & 25.2 & 33.5 & 31.9 & 31.2 & 29.9 & 30.7 &
30.3
\\
 & Whisper Medium &
30.6 & \textbf{39.0} & 25.2 & 35.3 & 37.5 & 29.5 & 29.5 & 31.1 &
32.2
\\
 & Whisper Large V2 &
31.7 & \underline{38.0} & 26.7 & 33.5 & 29.9 & \underline{31.8} & 28.7 & 30.8 &
31.4
\\

\hdashline

\multirow{4}{*}{Llama 2 13B} & Oracle & 21.5 & 30.0 & 18.5 & 18.4 & 25.7 & 20.8 & 27.7 & 32.1 & 24.3
\\

 & Whisper Small & \underline{35.8} & 35.0 & \underline{39.3} & \underline{35.7} & \underline{41.0} & 28.9 & 36.2 & \underline{34.0} & 35.7
\\

 & Whisper Medium & \textbf{37.7} & 36.0 & \textbf{45.2} & \textbf{39.0} & \textbf{44.4} & \textbf{32.4} & \textbf{37.4} & \textbf{34.1} & \textbf{38.3}
\\

 & Whisper Large V2 & 34.7 & \underline{38.0} & 37.0 & \textbf{39.0} & 39.6 & \textbf{32.4} & \underline{36.8} & 33.1 & \underline{36.3}
\\

\hline

\end{tabular}%
}
\vspace{-2mm}
\end{table*}













\begin{table}[H]
\caption{Audio duration distribution according to the labels.}
\label{tab:audio-distribution}
\centering
\resizebox{\columnwidth}{!}{%
\begin{tabular}{cccccc}
\hline
               & \textbf{MMLU} & \textbf{MedQA} & \textbf{MedMCQA} & \textbf{Total} & \textbf{\# Doc.} \\
\hline
\textbf{A}     & 1h50          & 5h55           & 5h41  & 13h28 & 1,936      \\
\textbf{B}     & 1h54          & 5h08           & 4h31  & 11h33 & 1,648    \\
\textbf{C}     & 1h50          & 5h49           & 3h57  & 11h37 & 1,519    \\
\textbf{D}     & 3h03          & 4h28           & 3h30  & 11h03 & 1,442    \\
\hline
\textbf{Total} & 8h39          & 21h22          & 17h40 & 47h41 & 6,545         \\
\hline
\end{tabular}
}
\end{table}






\subsection{Audio prompt format}
\label{s:prompt}

We standardized all textual MCQA datasets and synthesized them into audio format. These audio MCQAs serve as prompts for the studied and proposed SQA systems. Following experimentation with various formats and careful listening to the resulting audio outputs, we identified an effective format exemplified below:




\begin{figure}[!htb]
\centering
\begin{instructionframe}{Prompt Format}
{ \normalsize
\textit{A 39-year-old woman, with a history of thyroidectomy and primary hyperparathyroidism presents for surgical evaluation for a right adrenal mass.} \textbf{Preoperatively, which of the following medications should she receive to prevent a hypertensive emergency intraoperatively?} Option A: Atenolol \hspace{3mm} Option B: Labetolol Option C: Nifedipine \hspace{1mm} \textbf{Option D: Phenoxybenzamine} The correct answer is Option \colorbox{mybluelight}{\textbf{D}}







}
\end{instructionframe}
\label{prompt:prompt-format}
\end{figure}






\vspace{-8mm}

\subsection{Evaluation metric}
\label{s:eval}

The evaluation of multi-choice SQA with a single correct answer resembles a multi-class classification task. The performance is here assessed for each task using {\it Accuracy}, which measures the proportion of correctly predicted answers compared to the total number of questions. A prediction is considered accurate if it exactly matches the ground-truth answer, otherwise, it is classified as incorrect. Choosing the accuracy enables direct comparison with previous works on textual datasets \cite{nori2023capabilities, labrak2024biomistral}.

\section{Studied and proposed methods}
\label{sec:Method}


This section outlines the zero-shot approaches studied for SQA. Firstly, we introduce baseline models with cascade systems (Section~\ref{sec:cascade}). Then, we present models integrating our end-to-end audio-text entailment approach (Section~\ref{sec:zero-shot-entailement}). 


\subsection{Baseline cascade approaches}
\label{sec:cascade}

Our baseline models involve a two-stage process: transcription of audio inputs into text using an ASR module, followed by its processing with an LLM to select the correct answer to posed questions. We conducted experiments with various models to assess the impact of different ASR and LLM configurations on SQA performance. In the ASR stage, we compared the performance using the reference transcription ({\it Oracle}) against Whisper Small, Medium, and Large V2 ASR models to identify potential transcription error propagation issues. Subsequently, in the LLM stage, we compared the performance of an LLM similar in size to Whisper Large V2 (1.5 billion parameters), named Phi 1.5, against larger models based on the LLaMa 2 architecture, configured with 7B and 13B parameters, to assess the scalability of performance with model size. In total, we investigated 12 cascade system combinations.



During the second step of inference, the LLM predicts the next token based on the input prompt, generating probabilities for each token in the vocabulary. To ensure relevance, the vocabulary is filtered to include only relevant tokens (in this case, choice letters) corresponding to the expected answer options. This approach prevents the model from generating irrelevant tokens or hallucinations~\cite{liang2023holistic}. 






\subsection{Zero-shot end-to-end entailment-based approaches}
\label{sec:zero-shot-entailement}


Numerous studies~\cite{halder-etal-2020-task, pmlr-v37-romera-paredes15} have underscored the advantages of leveraging Natural Language Inference (NLI) for textual zero-shot entailment and classification tasks. However, except for CLAP~\cite{10095969} and Pengi~\cite{deshmukh2023pengi}, based on contrastive learning and prefix-tuning respectively, a limited adaptation of such methodologies has been observed in speech-related literature, particularly with large-scale pre-trained audio models like Whisper and SpeechGPT.
Our proposed zero-shot audio-text entailment method is integrated into the four previously mentioned models, aiming to assess the likelihood of a textual sequence matching an audio recording. In our setup, the audio contains the question and options, while the text represents classes A to D.



For Whisper~\cite{radford2022robust}, we utilize audio features and request individual log probabilities for each letter using the format:  \textit{<|startoftranscript|> [A] <|endoftext|>}.  The predicted class is determined by the highest average log probability. To comply with Whisper's 30-second limit for audio segments, we truncate segments beyond this duration to capture only the question and options. 
For SpeechGPT~\cite{zhang-etal-2023-speechgpt}, we populate the model's context in a prompt filled with speech units obtained from HuBERT~\cite{10.1109/TASLP.2021.3122291} representations discretized using k-means clustering on 1,000 clusters.
We then request the generation of one additional token to the model. Subsequently, we filter the vocabulary to retain only the log probabilities corresponding to letters A to D, as described earlier in Section~\ref{sec:cascade}. Pengi~\cite{deshmukh2023pengi} undergoes minimal changes in the model, audio representation, and prompt format, maintaining a similar procedure.  
The approach is slightly adapted for the CLAP model~\cite{10095969}, a dual encoder architecture trained with contrastive language-audio pre-training. Here, individual encoders process both speech and text. Given an audio sample ($a$) and a list of classes ($o$), we identify the best match among all pairs by calculating the cosine distance between their vector representations. The pair with the closest distance is considered the predicted match.



\begin{table*}[!t]
\centering

\setlength\extrarowheight{1.5pt}
\caption{Accuracy (in \%) of the zero-shot end-to-end models applying our entailment method. 
Highest value in bold and second best is underlined, excluding SpeechGPT + Oracle (model aligned with reference transcriptions).}
\label{tab:results-entailment}
\resizebox{\textwidth}{!}{%
\begin{tabular}{lllllllllll}

\hline

\multicolumn{1}{c}{\textbf{}} &
\multicolumn{1}{c}{\textbf{}} &
\multicolumn{6}{c}{\textbf{MMLU}} &
\multicolumn{1}{c}{\textbf{}} &
\multicolumn{1}{c}{\textbf{}} \\ \cline{2-8}

\multicolumn{1}{c}{\textbf{}} &
\multicolumn{1}{c}{\textbf{}} &
\multicolumn{1}{c}{\textbf{Clinical KG}} &
\multicolumn{1}{c}{\textbf{Medical Genetics}} &
\multicolumn{1}{c}{\textbf{Anatomy}} &
\multicolumn{1}{c}{\textbf{Pro Medicine}} &
\multicolumn{1}{c}{\textbf{College Biology}} &
\multicolumn{1}{c}{\textbf{College Medicine}} &
\multicolumn{1}{c}{\textbf{MedQA}} &
\multicolumn{1}{c}{\textbf{MedMCQA}} &
\multicolumn{1}{c}{\textbf{Avg.}} \\
\hline



\multirow{3}{*}{Whisper} & Small & 
24.1 & 
\textbf{31.0} & 
20.0 & 
17.6 & 
25.0 & 
20.2 & 
\textbf{27.7} & 
\underline{30.6} &
24.5 \\

& Medium &
\textbf{30.6} & 
20.0 & 
17.8 & 
\underline{42.6} & 
\underline{26.4} & 
30.6 & 
21.9 & 
22.5 &
26.5 \\

& Large V2 &
27.5 & 
24.0 & 
26.7 & 
20.2 & 
20.1 & 
19.6 & 
25.8 & 
27.4 &
23.9
\\

\hdashline

\multirow{3}{*}{CLAP} & Unfused &   26.8 & 23.0 & 24.4 & 37.1 & \textbf{29.2} & \underline{32.9} & 23.1 & 19.7 & \underline{27.0} \\


& Large General &  \underline{29.4} & 21.0 & 23.7 & \textbf{44.5} & 25.7 & \textbf{34.1} & 21.1 & 20.3 & \textbf{27.5} \\

& Fused & 21.5 & \underline{30.0} & 18.5 & 18.4 & 25.7 & 20.8 & \textbf{27.7} & \textbf{32.0} & 24.3 \\





\hdashline

\multirow{2}{*}{Pengi} & Base & 24.9 & 26.0 & \textbf{32.6} & 21.3 & 19.4 &  24.8 & 24.0 & 24.4 & 24.7 \\
 & Base No Text Encoder & 26.8 & 26.0 & 25.2 & 20.2 & 22.2 & 20.8 & 24.3 & 25.9 & 23.9 \\

\hdashline

\multirow{1}{*}{SpeechGPT} & E2E &
28.3  &
23.0  &
\underline{29.6} &
17.6 &
21.5 &
27.2 &
\underline{26.4} &
23.4 &
24.6
\\

\hline
\hline

\multirow{1}{*}{SpeechGPT} & Oracle &
36.2  &
32.0 &
27.4  &
35.7  &
29.9  &
34.1  &
24.4 &
27.2 &
30.8
\\ 

\hline
\end{tabular}%
}
\vspace{-2mm}
\end{table*}








\section{Results}

In this section, we examine the zero-shot condition performance on our SQA tasks using first the baseline cascade models (Section~\ref{sec:zero-shot-cascade}), and then our entailment approach across various end-to-end models (Section~\ref{sec:zero-shot-E2E}).

\subsection{Zero-shot cascade approaches}
\label{sec:zero-shot-cascade}



Table~\ref{tab:WER} outlines the transcription performance, measured in Word Error Rate (WER), of Whisper ASR versions (Small, Medium, and Large V2) across various SQA tasks. Generally, Whisper Large V2 shows improved WER performance, except in MMLU Anatomy, where Whisper Medium performs better.

\begin{table}[H]
\centering
\caption{Transcription performance (in WER) on each SQA task. 
Best result in bold and second best is underlined.}
\label{tab:WER}
\resizebox{0.85\columnwidth}{!}{%

\begin{tabular}{ccccc}

\hline

\multirow{2}{*}{} & \multirow{2}{*}{\textbf{Tasks}} & \multicolumn{3}{c}{\textbf{Whisper}} \\ \cline{3-5} 

& & \textbf{S} & \textbf{M} & \textbf{L-V2} \\ \hline

\multicolumn{1}{c|}{\multirow{6}{*}{\text{MMLU}}}  & \text{Clinical KG} & 5.45 & \underline{4.21} & \textbf{3.30} \\

\multicolumn{1}{c|}{} & \text{Medical Genetics} & 6.19 & \underline{4.59} & \textbf{4.31} \\
\multicolumn{1}{c|}{} & \text{Anatomy} &  4.90 & \textbf{2.68} & \underline{3.50} \\
\multicolumn{1}{c|}{} & \text{Pro Medicine} & 5.66 & \underline{4.68} & \textbf{4.54} \\
\multicolumn{1}{c|}{} & \text{College Biology} &  4.54 & \underline{2.91} & \textbf{2.66} \\
\multicolumn{1}{c|}{} & \text{College Medicine} & 26.02 & \underline{25.54} & \textbf{24.74} \\
& \text{MedQA} & 7.50 & \underline{6.21} & \textbf{5.84}\\
& \text{MedMCQA} & 7.99 & \underline{6.33} & \textbf{6.10} \\
\hline
\multicolumn{2}{c}{\text{Average}} & 8.53 &	\underline{7.14}	& \textbf{6.87} \\
\hline
\end{tabular}%
}
\vspace{-1mm}
\end{table}

Table~\ref{tab:results-cascade} displays the accuracy performance of studied LLM-based zero-shot cascade methods using Whisper automatic transcriptions on multiple SQA tasks. Interestingly, the Whisper model with the lowest WER might not always be the optimal choice in a cascade approach, indicating a lack of direct correlation between WER and SQA accuracy. Conversely, SQA performance appears to depend on LLM size, with larger models yielding higher accuracy. Notably, there is an 11.67\% difference between Phi 1.5 and LLaMa 2 13B in Whisper Medium results, highlighting the significant advantage of scaling up LLMs.
Except for Phi 1.5, all models show improved performance with transcriptions compared to Oracle. This enhancement, particularly in LLaMa 2 architectures, may be attributed to their better adaptability to speech normalization formats, reduced punctuation, and increased noise.





Furthermore, with LLaMa 2, Whisper Medium transcriptions emerge as the top performers. Notably, LLaMa 13B demonstrates a 1.95\% overall accuracy gain over Whisper Large V2 and a 2.54\% improvement over Whisper Small. Similar trends are observed in the 7B model, with increases of 0.8\% over Large V2 and 1.9\% over Small. The performance of the LLaMa 2 13B model in a zero-shot scenario with Whisper Medium transcriptions shows promising results.

\subsection{Zero-shot end-to-end models' capabilities}
\label{sec:zero-shot-E2E}



Table~\ref{tab:results-entailment} outlines the accuracy performance of zero-shot end-to-end models using our entailment method on our multiple-choice SQA benchmark. While the overall average accuracy remains similar across models, specific models demonstrate proficiency in particular tasks, with none consistently outperforming others across all tasks. Notably, Whisper Medium showcases competitive zero-shot performance, surpassing cascade setups with Phi 1.5 despite having approximately half the parameters. CLAP's contrastive modeling outperforms Phi 1.5 but falls short of LLaMa 2 7B. Impressively, despite its smaller size—153M parameters in its base form and 193M in its larger form—CLAP performs remarkably well, being 14.7 times smaller than Whisper Large V2 combined with Phi 1.5 and 44.3 times smaller with LLaMa 2 7B. SpeechGPT encounters challenges in zero-shot tasks from speech, contrasting its performance with text (Oracle), highlighting difficulties in directly handling speech modality representations, which need to be addressed in the future, with a better alignment approach. Notably, Whisper, especially Whisper Medium, occasionally outperforms cascade configurations with Phi 1.5 in zero-shot scenarios. Specific tasks exhibit varying levels of difficulty for different models; for instance, MedMCQA yields high results with Whisper Small and CLAP Fused, while MMLU College Medicine favors Whisper Medium, CLAP Unfused, and CLAP Large General. SpeechGPT generally underperforms across most tasks, except for MMLU Anatomy and MedQA, where it outperforms most other models. Despite the small performance improvement over cascade systems, which is linked to the zero-shot setting, E2E systems can be enhanced by scaling with better quality SQA data and increasing the number of parameters to see if they follow scaling laws similar to LLMs.

\section{Analysis of encoder layers}

This section presents an extensive analysis to pinpoint the critical location of information crucial for SQA tasks within the layers encoding the audio signal. To conduct this analysis, we extracted a subset of the MedMCQA training set consisting solely of audio sequences shorter than 30 seconds, which comprised 97.56\% of the data, resulting in 120 hours of spoken data. This subset was partitioned into training and validation sets using an 80\%/20\% ratio, yielding 95 hours and 23 hours, respectively. Our experimental approach involves fine-tuning audio encoders and introducing an intermediate trainable layer of equal size to the number of encoder layers. This intermediate layer selects information from the encoder's layers through a weighted sum of their representations when feeding the classification head. The objective of this weighted encoder layers approach is to analyze the necessity of specific layers for executing the SQA task while enhancing model understanding.

\begin{figure}[H]
  \centering
  \includegraphics[width=0.95\linewidth]{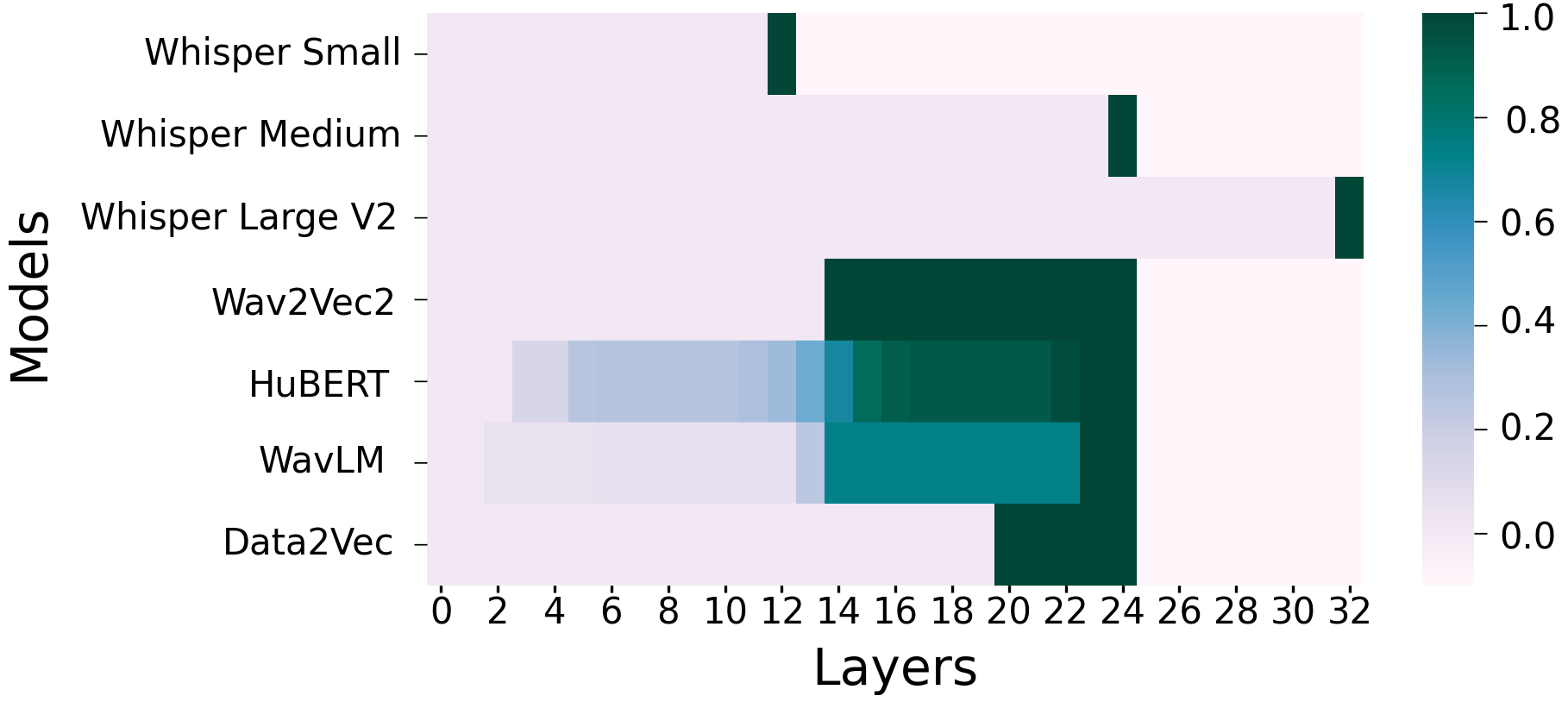}
  \caption{Cumulative weights according to encoder layers.}
  \label{fig:CummulativeWeight}
\end{figure}

As depicted in Figure~\ref{fig:CummulativeWeight}, illustrating cumulative weights across encoder layers, Whisper models exhibit a propensity to concentrate information in the final layers, aligning with prior research findings~\cite{yang23d_interspeech}. This indicates that these audio-based models effectively utilize the last layer to represent textual information, possibly due to heavy reliance on the decoder.

In contrast, Wav2Vec~\cite{10.5555/3495724.3496768} and Data2Vec~\cite{pmlr-v162-baevski22a} primarily rely on a single intermediate layer, specifically the 15th and 21st layers, respectively. However, HuBERT~\cite{10.1109/TASLP.2021.3122291} and WavLM~\cite{9814838} adopt a different strategy,  integrating information from a broader range of layers. HuBERT integrates data from 12 layers, while WavLM incorporates information from 4 layers distributed across various regions of the encoder.







\section{Conclusion}


This study introduces a novel synthetic Spoken Question Answering (SQA) dataset tailored specifically to the medical domain. We conducted zero-shot comparative analyses of end-to-end speech methodologies using a new entailment technique against cascade speech transcription and LLM module. Our experiments and analysis demonstrate the effectiveness of our end-to-end approach, yielding performances comparable to those achieved by cascade models of similar sizes. Moving forward, we aim to explore the utilization of speech alignment techniques with LLMs to enhance end-to-end question answering performance, with a particular emphasis on improving outcomes in low-resource domains such as healthcare. Our research faced multiple constraints. Using limited speaker variety for synthetic audio may reduce accuracy compared to natural speech, affecting response precision. Simplifying task formulation lacks genuine human interaction dynamics but enables metric-based assessments, enhancing model reproducibility and cost efficiency. Finally, our study neglects multilingual contexts, highlighting the need for additional exploration in diverse linguistic settings.

\section{Acknowledgements}

This work was performed using HPC resources from GENCI-IDRIS (Grant 2024-AD011015344 and Grant 2022-AD011013061R2). This work was financially supported by ANR MALADES (ANR-23-IAS1-0005) and Zenidoc.




\bibliographystyle{IEEEtran}
\bibliography{mybib}

\end{document}